\documentclass[journal]{IEEEtran}

\usepackage{amsmath,amsfonts}
\usepackage{algorithmic}
\usepackage{array}
\usepackage[caption=false,font=normalsize,labelfont=sf,textfont=sf]{subfig}
\usepackage{textcomp}
\usepackage{stfloats}
\usepackage{url}
\usepackage{verbatim}
\usepackage{graphicx}

\usepackage{cite}
\usepackage{multirow}
\usepackage[pagebackref=true,breaklinks=true,colorlinks,bookmarks=false]{hyperref}
\usepackage{bm}
\usepackage{booktabs}

\def\eg{\emph{e.g.}}

\def\BibTeX{{\rm B\kern-.05em{\sc i\kern-.025em b}\kern-.08em
    T\kern-.1667em\lower.7ex\hbox{E}\kern-.125emX}}
\usepackage{balance}

\begin{document}

\title{Holistic Prototype Attention Network for Few-Shot VOS}

\author{
	Yin~Tang*,
	Tao~Chen*,
	Xiruo~Jiang,
	Yazhou~Yao,
	Guo-Sen~Xie
	and~Heng-Tao~Shen (Fellow, IEEE)
	\thanks{Yin~Tang, Tao~Chen, Yazhou~Yao, Guo-Sen~Xie and Xiruo~Jiang are with the School of Computer Science and Engineering, Nanjing University of Science and Technology, Nanjing, China.}
	\thanks{Heng-Tao~Shen is with the School of Computer Science and Engineering, University of Electronic Science and Technology of China, Chengdu, China.}
    \thanks{*Equal contribution.}
}

\markboth{}%
{Holistic Prototype Attention Network for Few-shot Video Object Segmentation}

\maketitle

\begin{abstract}
	Few-shot video object segmentation (FSVOS) aims to segment dynamic objects of unseen classes by resorting to a small set of support images that contain pixel-level object annotations. Existing methods have demonstrated that the domain agent-based attention mechanism is effective in FSVOS by learning the correlation between support images and query frames. However, the agent frame contains redundant pixel information and background noise, resulting in inferior segmentation performance. Moreover, existing methods tend to ignore inter-frame correlations in query videos. To alleviate the above dilemma, we propose a holistic prototype attention network (HPAN) for advancing FSVOS. Specifically, HPAN introduces a prototype graph attention module (PGAM) and a bidirectional prototype attention module (BPAM), transferring informative knowledge from seen to unseen classes. PGAM generates local prototypes from all foreground features and then utilizes their internal correlations to enhance the representation of the holistic prototypes. BPAM exploits the holistic information from support images and video frames by fusing co-attention and self-attention to achieve support-query semantic consistency and inner-frame temporal consistency. Extensive experiments on YouTube-FSVOS have been provided to demonstrate the effectiveness and superiority of our proposed HPAN method. Our source code and models are available anonymously at \url{https://github.com/NUST-Machine-Intelligence-Laboratory/HPAN}.
\end{abstract}


\section{Introduction}
\label{sec:introduction}

\IEEEPARstart{W}{ith} the rapid advancements of CNN-based~\cite{He2016DeepRL} and attention-based~\cite{Vaswani2017AttentionIA} models, neural networks have achieved significant performance boosts in various vision tasks, such as object detection~\cite{liu2020video,wang2019revisiting,yao2023automatedor}, captioning~\cite{yan2022Videocaptioning,Wu2022TowardsKV,xu2019dualstreamrn} and semantic segmentation~\cite{liu2021sg,zhou2022survey,lu2020learning, chen2023multigranularityda, yao2021nonsalientro}. Among all vision tasks, video object segmentation (VOS) is a task that aims to segment objects in video frames. It has attracted increasing attention from researchers in recent years. VOS consists of the following two subtasks: (1) unsupervised VOS (UVOS) segments salient objects without clues and (2) semi-supervised VOS (SVOS) uses the first frame mask to segment specific entities. However, existing VOS models rely on large amounts of training data. Moreover, these models can hardly recognize unseen classes. Consequently, few-shot video object segmentation (FSVOS), which uses support images to help discover objects of unseen classes in query videos, is proposed to address these issues.

\begin{figure}[t]
	\centering
	\includegraphics[width=1\columnwidth]{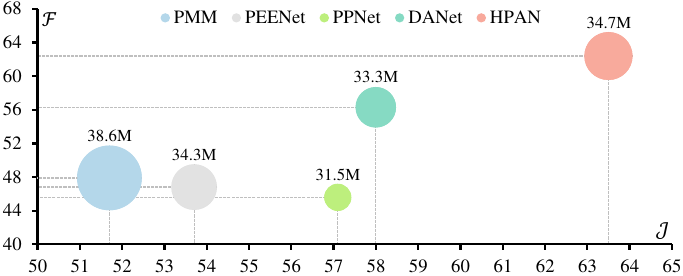}
	\caption{Comparison between existing FSVOS methods and our HPAN on the YouTube-FSVOS dataset. The x-axis and y-axis represent the average of the intersection over union $\mathcal{J}$ and the contour accuracy $\mathcal{F}$, respectively. The circle size indicates the model size in each method.}
	\label{fig:motivation}
\end{figure}

The FSVOS method differs from the existing VOS tasks relying on human intervention or online labeling. It aims to accurately and quickly segment unseen objects in videos by flexibly using a few static annotated support images.
In contrast to few-shot semantic segmentation (FSS), FSVOS concentrates on real-world videos with dynamic frames and is suitable for application scenarios such as autonomous driving and video surveillance. Although a new setting, FSVOS with support set flexibility and query set dynamics is indeed an important task setting.
Existing FSVOS methods can be broadly categorized into two types: methods that use temporal transduction inference and methods that use domain agent attention. TTI~\cite{Siam2022TemporalTI} proposes to employ temporal transduction inference to train a linear classifier for each frame by a class-consistent global loss and a foreground/background proportion-consistent local loss. However, TTI~\cite{Siam2022TemporalTI} requires online training for each new query frame, which can be computationally expensive and slow. This hinders the application of TTI in real-time video segmentation scenarios. DANet~\cite{Chen2021DelvingDI} adopts a meta-learning-based domain agent network to decompose many-to-many full-rank attention into two small attention blocks by extracting the middle frame as a domain agent. However, the domain agent attention has the following disadvantages. First, extracting a complete frame as a domain agent brings redundant features of foreground objects and background noise. This leads to high computational costs and inaccurate focus of the attention module on the target object. Second, DANet~\cite{Chen2021DelvingDI} only selects one frame as the agent, which may cause a loss of holistic information among all support images and query frames. Lastly, the support-to-query co-attention neglects the temporal consistency in query videos. In this work, we also employ the meta-learning-based paradigm to avoid tuning network weights in the inference period.

To address the aforementioned issues of meta-learning-based methods, we propose a holistic prototype attention network (HPAN) to fully leverage holistic information from support images and query frames. Our approach departs from the existing method DANet~\cite{Chen2021DelvingDI}, which rely on the middle frame as the domain agent to simplify multi-frame attention. HPAN replaces the domain agent with representative prototypes extracted from all foreground features. By leveraging these prototypes, HPAN is able to capture the overall relationship among all the supports and the queries, and achieve multi-frame co-attention and self-attention in a more efficient manner. HPAN mainly consists of two modules, a prototype graph attention module (PGAM) and a bidirectional prototype attention module (BPAM). PGAM is proposed to suppress the background noise of features and obtain overall representative prototypes by clustering and integrating all foreground local features. It generates pseudo-masks for query frames to filter out the background noise from the third-layer features. Then, it clusters local prototypes from foreground features using the k-means algorithm. Subsequently, PGAM fuses two groups of raw prototypes with two graph attentions and one graph co-attention, capturing the inner relationship among all prototypes. BPAM is designed to explore integrative attention, ensuring inner-frame temporal consistency and support-query semantic consistency. It introduces a prototype self-attention and a prototype co-attention to simplify full-rank attention and leverage temporal information among all query frames. Moreover, we have introduced a prototype loss to prevent the over-concentration of prototypes.

To our knowledge, this is the first attempt in the field of FSVOS to leverage representative prototypes for capturing the overall relationship between support images and query frames. Our proposed method has displayed remarkable efficiency on the YouTube-FSVOS dataset, illustrated in Fig.~\ref{fig:motivation} for comparative performance and computational cost analysis. The contributions of this paper are outlined as follows:

\begin{itemize}
	\item We propose a new framework, called holistic prototype attention network (HPAN), to improve the FSVOS
	      performance by fully leveraging holistic information.
	\item We construct a prototype graph attention module (PGAM) to obtain representative holistic prototypes by leveraging the inner correlations among all support images and query frames.
	\item  We design a bidirectional prototype attention module (BPAM), integrating co-attention and self-attention, to exploit temporal information in query frames by prototype-transductive reasoning.
	\item Extensive experimental results on YouTube-FSVOS show that our HPAN surpasses the state-of-the-art performance. Thorough ablation studies are conducted to verify the effectiveness of our method.
\end{itemize}

This paper is structured as follows: in Section~\ref{sec:related_works}, we present a comprehensive review of related studies; Section~\ref{sec:approach} outlines our proposed approach, while Section~\ref{sec:experiments} presents our experimental evaluations and ablation studies on the YouTube VOS dataset for the task of few-shot video object segmentation. Finally, we conclude our work in Section~\ref{sec:conclusion}.

\section{Related Works}
\label{sec:related_works}

\subsection{Video Object Segmentation}
The current video object segmentation (VOS) tasks include unsupervised VOS and semi-supervised VOS. Unsupervised VOS~\cite{Perazzi2016ABD,Wang_2019_ICCV,Wang_2019_CVPR,pei2023hierarchicalcp,zhou2021target} aims to segment conspicuous targets without the need for annotation. Various UVOS methods have been proposed based on several techniques, such as two-stream networks~\cite{Zhou2020MatnetMT, Ren2021ReciprocalTF, Pei2022HierarchicalFA}, inner-frame propagation~\cite{Fan2022BidirectionallyLD, Garg2021MaskSA}, and space-temporal memory~\cite{Liu2022GlobalSF, Lai2020MastAM}. For example, MATNet~\cite{Zhou2020MatnetMT} proposed a two-stream motion-attentive transition network to leverage motion cues as a bottom-up signal to guide the perception of object appearance. DBSNet~\cite{Fan2022BidirectionallyLD} adopted a dense bidirectional Spatio-temporal feature propagation network to integrate the forward and the backward propagation. GSFM~\cite{Liu2022GlobalSF} introduced a global spectral filter memory network, fusing the low-frequency feature of past frames in a space-time memory block and enhancing the high-frequency feature in a detail-aware decoder. Nevertheless, UVOS methods are unable to segment pre-specified target objects due to a lack of pixel-level annotation. Semi-supervised VOS (SVOS)~\cite{Caelles2017OneShotVO,lu2020video,wang2018semi} endeavors to achieve dense pixel tracking on target objects when the first frame mask is given. Existing SVOS works primarily employ propagation-based~\cite{Yang2021AssociatingOW} or matching-based~\cite{Cheng2021RethinkingSN} strategies. For instance, LSTT~\cite{Yang2021AssociatingOW} proposed a long short-term transformer to merge the long-term and short-term propagation features. STCN~\cite{Cheng2021RethinkingSN} adopted the L2 similarity to compute spatio-temporal affinity between two frames, effectively guiding segmentation heads. However, the first frame annotation is hard and expensive to acquire in practical scenarios. Moreover, many memory-based~\cite{Liang2021VideoOS, Mao2021JointIA, Seong2021HierarchicalMM, Park2022PerclipVO, Lin2022SwemTR} approaches have emerged recently to store previous frame features for subsequent frame segmentation.

\begin{figure*}[!t]
	\centering
	\includegraphics[width=1\linewidth]{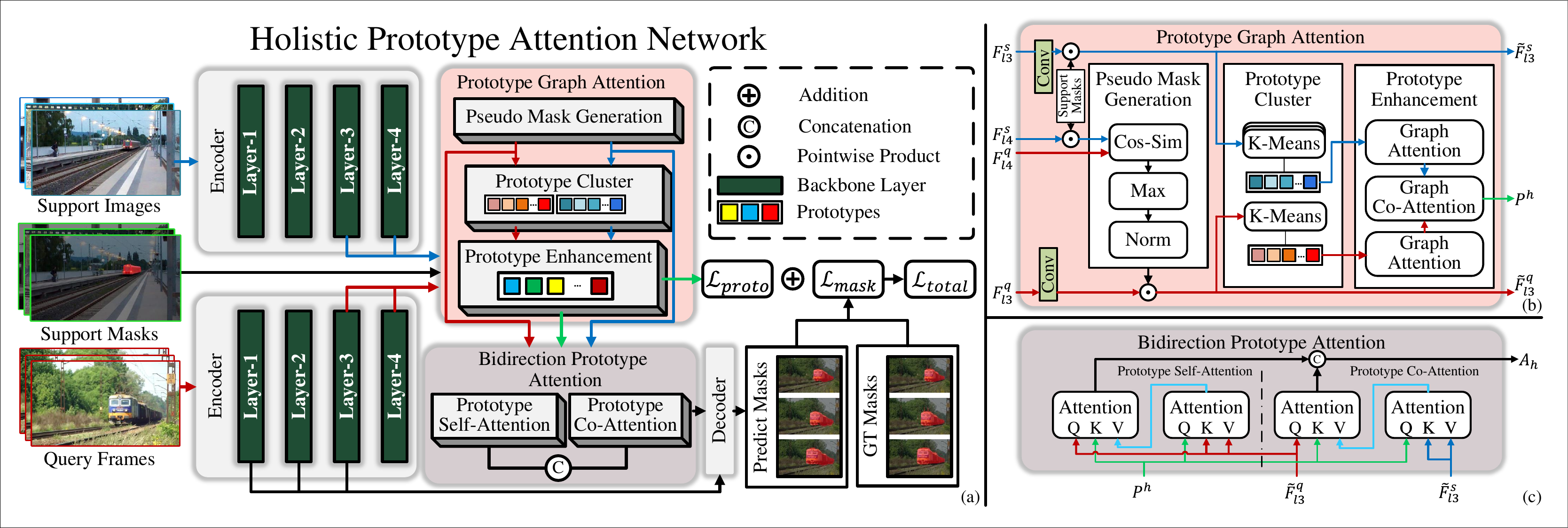}
	\caption{
		Overall architecture of our proposed HPAN.
		(a) Pipeline of HPAN.
		Support images and query frames are fed to the encoder to obtain four-level features.
		Third- and fourth-layer features are sent into the PGAM with support masks to obtain foreground features and holistic prototypes.
		The output of PGAM is then passed into BPAM for computing holistic attention.
		Finally, the holistic attention and the first three layers of query features are input to the decoder for predicting masks.
		(b) Pipeline of PGAM.
		PGAM generates pseudo-masks to suppress the background features and integrates all foreground features into the holistic prototypes by k-means and graph attention.
		(c) Pipeline of BPAM.
		BPAM concatenates the prototype self-attention and prototype co-attention for obtaining holistic attention.
        }
	\label{fig:pipeline}
\end{figure*}

\subsection{Prototype Learning}
The prototype learning techniques has been recognized across various domains, including image classification~\cite{Yang2018RobustCW, wang2023visualrw} and semantic segmentation~\cite{wang2023visualrw, Zhou2022RethinkingSS}. Deep Nearest Centroid (DNC)~\cite{wang2023visualrw} employs the Sinkhorn-Knopp~\cite{knight2008thesa} algorithm to accelerate the clustering of features into various category sub-centroids. This technique replaces the softmax classification layer, enhancing deep network parameter optimization and significantly improving image classification and semantic segmentation tasks. Similarly, ProtoSeg~\cite{Zhou2022RethinkingSS} introduces an internal online clustering approach that quickly generates non-learnable prototypes from training data for metric learning, facilitating prototype-based pixel classification.
Furthermore, prototype learning has yielded advancements in label-limited tasks such as few-shot learning (FSL)~\cite{liu2020prototyperf, liu2019prototypepn} and zero-shot learning (ZSL)~\cite{xu2020attributepn, fu2017zeroshotlo}. The prototype rectification method~\cite{liu2020prototyperf} is proposed to minimize intra-class and inter-class bias in transductive learning via label propagation and feature transfer. In the realm of weakly-supervised FSL, Prototype Propagation Network (PPN)~\cite{liu2019prototypepn} propagates coarse-level class prototypes to fine-level prototypes by the attention mechanism. Attribute prototype network~\cite{xu2020attributepn} is employed to jointly learn global and local features, regress attributes from features, and decorrelate prototypes.

\subsection{Few-shot Semantic Segmentation}
Few-shot semantic segmentation (FSS)~\cite{Shaban2017OneshotLF} aims to perform image semantic segmentation using a few unseen class samples that contain annotations. Existing FSS approaches follow the metric learning framework, including parameter-based, prototype-based, and hybrid methods. The parameter-based methods~\cite{Lu2021SimplerIB, liu2023fecanetbf, Johnander2022DenseGP, Min2021HypercorrelationSF, Hong2022CostAW} compare the pairwise distance between query and support by a parameter-model, such as linear classification~\cite{Lu2021SimplerIB}, 4D-convolution~\cite{Min2021HypercorrelationSF}, and gaussian processes~\cite{Johnander2022DenseGP}. CWT~\cite{Lu2021SimplerIB} designed a classifier weight transformer to tune the weights of the transformer online with a support-set trained linear classifier which simplifies the meta-learning task. HSNet~\cite{Min2021HypercorrelationSF} introduced 4D-convolution to FSS to compress high-level semantic and low-level geometric cues. Hong~\cite{Hong2022CostAW} extended HSNet to 4D-transformer to enlarge the receptive field. DGPNet~\cite{Johnander2022DenseGP} proposed to fit a set of gaussian process models based on support images and their corresponding masks, to compute the mean and variance of the foreground feature distribution. The prototype-based approaches~\cite{Gao2022DrnetDR,Li2021AdaptivePL,Liu2020PartawarePN,Wang2019PanetFI,Yang2020PrototypeMM,liu2022intermediate} cluster the features into one or more prototypes representing particular semantics and compare prototypes with query features to discover the target area. PANet~\cite{Wang2019PanetFI} presented a prototype alignment regularization between support and query for class-specific knowledge representation to achieve better generalization performance. PPNet~\cite{Liu2020PartawarePN} proposed a prototype representation method that decomposed the overall class representation into partial prototype awareness to capture enriched and fine-grained feature representations. DRNet~\cite{Gao2022DrnetDR} aims to solve the intra-class variance of the unseen class through two recalibration modules, which explore latent regions for query images. IMPT~\cite{liu2022intermediate} suggests using an intermediate prototype to extract class information from deterministic support features and adaptive query features. Combining the above two types of methods, the hybrid methods~\cite{Lang2022LearningWN, Xie2021ScaleawareGN, Tian2020PriorGF} tend to obtain a robust performance. For example, PFENet~\cite{Tian2020PriorGF} proposed to concatenate the query features, support prototypes, and prior masks and send them into the feature enrichment module to improve model performance. BAM~\cite{Lang2022LearningWN} applied an additional learner to prompt the object region of base classes that do not need to be segmented.

Inspired by prototype-based methods that can effectively compress semantic information of objects, we propose a prototype graph attention module in this work. Unlike existing methods, we adopt two graph attention blocks and one graph co-attention to integrate the holistic prototypes.

\section{The Proposed Approach}
\label{sec:approach}

\subsection{Preliminaries}
FSVOS aims to train a model using data from base classes and then segment objects from unseen classes in a query video using a few labeled images. Following DANet~\cite{Chen2021DelvingDI}, we adopt a task setting including meta-learning and fine-tuning phases. Specifically, all videos in the dataset are divided into two sets: a training set $\mathcal{D}_{train}$ composed of base classes and a test set $\mathcal{D}_{test}$ comprised of unseen classes. Both $\mathcal{D}_{train}$ and $\mathcal{D}_{test}$ contain a series of extracted episodes, each of which consists of a support set $\mathcal{Q}$ and a query set $\mathcal{S}$, respectively. Each episode samples $K$ support images and $T$ query frames from the same class $c$. The query set is denoted as $\mathcal{Q}={\{(I^q_n,M^q_{n,c})\}}^T_{n=1}$, where $I^q_n$ is the $n$-th continuous frame from the selected query video and $M^q_{n,c}$ is the query mask of class $c$ corresponding to the $n$-th frame. Similarly, the support set is denoted as $\mathcal{S}_c= {\{(I^s_n,M^s_{n,c})\}}^K_{n=1}$, where $I^s_n$ is the $n$-th support image from different videos and $M^s_{n,c}$ is the support mask of class $c$ in the $n$-th image.

\subsection{Architecture Overview}
To improve the performance of few-shot video object segmentation, we propose a holistic prototype attention network that fully utilizes features of support and query sets. As shown in Fig.~\ref{fig:pipeline}, HPAN introduces a prototype graph attention module (PGAM) and a bidirectional prototype attention module (BPAM) to discover target regions more precisely. The proposed PGAM~(see Fig.~\ref{fig:pipeline}(b)) enhances the representation of prototypes and extracts latent local features of unseen objects. It generates pseudo-masks on query frames by comparing support and query features, which are used to separate latent foreground features. After clustering prototypes from the foreground features, we aggregate these prototypes through a graph attention network to facilitate the local details of the segmentation. The BPAM~(see Fig.~\ref{fig:pipeline}(c)) is designed to guarantee the external semantic consistency between supports and queries and the internal temporal consistency among all query frames. It concatenates the support-query co-attention with the query-query self-attention in a transductive learning manner. Moreover, we propose a prototype loss to prevent the prototype distribution from being concentrated and losing local details.

\subsection{Encoder and Decoder}
The encoder network $\mathcal{E}$ extracts features from the support image $I^s$ and the query frame $I^q$ as follows
\begin{equation}
	\begin{aligned}
		F^s_{l1},F^s_{l2},F^s_{l3},F^s_{l4}=\mathcal{E}(I^s) , \\
		F^q_{l1},F^q_{l2},F^q_{l3},F^q_{l4}=\mathcal{E}(I^q) .
	\end{aligned}
	\label{eq:encoder}
\end{equation}
$F^s_{li}$ and $F^q_{li}$ represent the $i$-th layer feature of support images and query frames, respectively. The backbones for the support set and the query set share weights. Different from DANet\cite{Chen2021DelvingDI}, we keep the fourth-layer features and feed the third- and fourth-layer features into PGAM to compute foreground ones and holistic prototypes. Besides, the first three-layer features of query frames are also fed into the decoder net $\mathcal{D}$ to refine the segmentation results as follows:
\begin{equation}
	\hat{M}^q=\mathcal{D}(A_h,F^q_{l1},F^q_{l2},F^q_{l3}),
	\label{eq:decoder}
\end{equation}
where the feature $A_h$ is the holistic attention from BPAM and $\hat{M}^q$ is the predicted masks of query frames.

\subsection{Prototype Graph Attention Module}
PGAM~(Fig.~\ref{fig:pipeline}(b)) is designed to suppress background noise when extracting features and generate prototypes for unseen objects in all support images and query frames. Unlike kMaX~\cite{yu2022kmeansmt} and CLUSTERSEG~\cite{liang2023clustsegcf}, which use cross-attention clustering to distinguish different instances or categories, our PGAM aims to extract local information of unseen objects using a cluster-then-enhancement method. Specifically, we first compute pseudo-masks for query frames by measuring distances between features. Then, we cluster foreground features based on a parameter-free k-means algorithm to aggregate prototypes. Finally, graph attention efficiently enhances query prototypes by integrating support information.

\textbf{Pseudo-mask Generation.}
Inspired by PFENet\cite{Tian2020PriorGF}, our method focuses on foreground features to suppress background noise. We first compute the similarity between each location in the query feature and its nearest support foreground feature. By further normalizing the similarity matrix, we obtain the pseudo-mask for the query feature at the fourth layer. The pseudo-masks are computed on fourth-layer features because deeper features have more semantic information and can help discover latent areas of unseen objects. Moreover, the shape of the fourth-layer features is smaller than other layers, which enables a lower computation cost when calculating the similarity.The fourth-layer foreground feature $\tilde{F}^S_{l4}$ is given as

\begin{equation}
	\tilde{F}^s_{l4} = F^s_{l4} \cdot M^s,
	\label{eq:foreground_q_l4}
\end{equation}
in which, $M^s$ denotes the mask of support images. The cosine similarity matrix is computed between each pixel of $F^q_{l4}$ and its closest pixel in $\tilde{F}^s_{l4}$. Every element of the similarity matrix denoted as $\overline{M}^q_{j}$ can be expressed in the following manner:
\begin{equation}
	\begin{aligned}
		\overline{m}^q_{j,t} = \max_{i,k}{d(\tilde{f}^s_{l4,i,k},f^q_{l4,j,t})}, \\
		i,j\in\{1,\cdots,H_{l4}W_{l4}\},
	\end{aligned}
	\label{eq:pseudo_mask_1}
\end{equation}
where $\tilde{f}^s_{l4,i,k}$ and $f^q_{l4,j,t}$ is the corresponding pixel of the $k$-th image feature $\tilde{F}^s_{l4}$ and the $t$-th frame feature $F^q_{l4}$, respectively. $H_{l4}$ and $W_{l4}$ are the height and width of the fourth layer feature. Cosine similarity is used to measure the distance between feature vectors $s$ and $t$ as follows:
\begin{equation}
	d(s,t)=\frac{<s,t>}{\parallel s\parallel\cdot\parallel t\parallel+\epsilon} ,
	\label{eq:cs}
\end{equation}
where $\epsilon$ is set to $10^{-8}$ to avoid division by zero. Once computed, we normalize the similarity matrix to generate pseudo masks as follows
\begin{equation}
	\tilde{m}^q_{j,t} = \frac{\overline{m}^q_{j,t} - \min_j{\overline{m}^q_{j,t}}}{\max_j{\overline{m}^q_{j,t}}-\min_j{\overline{m}^q_{j,t}}},
	\label{eq:pseudo_mask_2}
\end{equation}
where $\tilde{m}^q_{j,t}$ is the corresponding pixel of the $t$-th frame pseudo-mask $\tilde{M}^q$.

\textbf{Prototype Clustering.}
The goal of prototype clustering is to obtain support and query prototypes. In our implementation, we opt for k-means as our clustering algorithm based on three reasons below: (1)~k-means offers an intuitive and structurally straightforward clustering method that eliminates the necessity to learn additional model parameters; (2) it exhibits rapidity and efficiency in generating initial prototypes for unseen objects; and (3) the quantity of cluster centers in k-means is adjustable, thus preventing the under-representing or over-fitting of unseen target features due to an insufficiency or excess of prototypes.

As shown in Fig.~\ref{fig:pipeline}(b), we first perform a 1x1 convolution operation on the third-layer features to reduce the feature dimension, and then filter the background noise through support mask $M^s$ and pseudo-mask $\tilde{M}^q$, respectively:
\begin{equation}
	\begin{aligned}
		\tilde{F}^s_{l3} & = {\rm conv}(F^s_{l3})\cdot M^s ,         \\
		\tilde{F}^q_{l3} & = {\rm conv}(F^q_{l3})\cdot \tilde{M}^q .
		\label{eq:foreground_l3}
	\end{aligned}
\end{equation}
The number of prototypes in each image is given as $N_p$. We employ k-means to cluster the foreground features of each support image into $N_p$ clusters:
\begin{equation}
	P^s = \{{\rm kmeans}(\tilde{F}^s_{l3,k})|k=1,\cdots,K\},
	\label{eq:proto_s_raw_1}
\end{equation}
where $P^s\in \mathbf{R}^{(N_pK)\times C}$ is the raw support prototypes set and $K$ denotes the number of support images. Compared with the support images, there are more similar features in different frames of the query video. Therefore, we cluster the foreground features of all query frames into $N_pT$ clusters to promote the variety of query prototypes:
\begin{equation}
	P^q = {\rm kmeans}(\tilde{F}^q_{l3}) ,
	\label{eq:proto_q_raw}
\end{equation}
where $P^q\in \mathbb{R}^{(N_pT)\times C}$ is the raw query prototypes set.

\begin{figure}[!t]
	\centering
	\includegraphics[width=1\columnwidth]{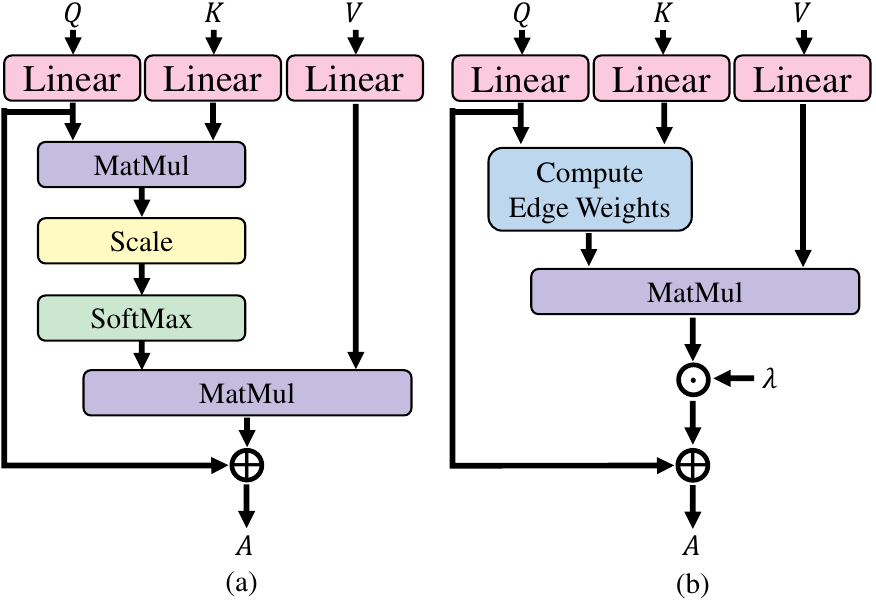}
	\caption{The structures of (a) Common Attention $\mathcal{A}$ and (b) Graph Attention $\mathcal{G}$. $\bigoplus$: addition, $\bigodot$: pointwise product}
	\label{fig:graph_attn}
\end{figure}

\textbf{Prototype Enhancement.}
Inspired by PPNet\cite{Liu2020PartawarePN}, we propose a graph attention module to improve the representation ability of prototypes. We enhance support and query prototypes using two graph self-attention blocks and then fuse all prototypes in a graph co-attention module. Compared to the existing feature enhancement module in TF-Blender~\cite{cui2021tfblendertf}, our method improves the local prototypes using graph attention modules to better capture the relationship between support and query. The computation of graph attention $\mathcal{G}$ is shown in Fig.~\ref{fig:graph_attn} (b).
\begin{equation}
	\begin{aligned}
		\mathcal{G}(P^{tgt},P^{src},\lambda) & = P^{query}+\lambda \Phi P^{value} , \\
		P^{key}                              & = W_kP^{src},                        \\
		P^{query}                            & = W_qP^{tgt},                        \\
		P^{value}                            & = W_vP^{tgt} ,
	\end{aligned}
	\label{eq:ga}
\end{equation}
where $W_{k}$, $W_{q}$, $W_{v}$ are the linear parameters of the graph attention block. $\lambda$ is a scaling coefficient and $\Phi$ is the edge weights between source prototypes $P^{key}$ and target prototypes $P^{query}$. Each element of edge weights $\Phi$ is
\begin{equation}
	\phi_{ij} = \frac{d(p^{key}_{i},p^{query}_{j})}{\sum_{j'=1}^{N_{query}}d(p^{key}_{i},p^{query}_{j'})} ,
	\label{eq:ga_weight}
\end{equation}
where $d(\cdot,\cdot)$ is the cosine similarity (Eq.~\ref{eq:cs}), $p^{key}_{i}$ and $p^{query}_{j}$ is the corresponding prototype in $P^{key}$ and $P^{query}$, respectively. The three graph attention operations in PGAM are denoted as follows:
\begin{equation}
	\begin{aligned}
		\overline{P}^s & = \mathcal{G}(P^s,P^s,\lambda_{self}) ,                     \\
		\overline{P}^q & = \mathcal{G}(P^q,P^q,\lambda_{self}) ,                     \\
		P^h            & = \mathcal{G}(\overline{P}^s,\overline{P}^q,\lambda_{co}) ,
	\end{aligned}
	\label{eq:qga}
\end{equation}
where $P^h\in \mathbb{R}^{(N_pK)\times C}$ is the holistic prototypes, $\lambda_{self}$ and $\lambda_{co}$ are the corresponding coefficient of graph self-attention and graph co-attention.

\subsection{Bidirectional Prototype Attention Module}
To guarantee the internal temporal consistency among all query frames and the external semantic consistency between supports and queries, we further propose the bidirectional prototype attention module (BPAM). As shown in Fig.~\ref{fig:pipeline}(c), we integrate the support-query co-attention and the query-query self-attention for maintaining these consistencies by prototype transductive reasoning manner. Here, we adopt the commonly-used attention $\mathcal{A}$ with skip connections (Fig.~\ref{fig:graph_attn} (a)) as the basic block in BPAM:
\begin{equation}
	\begin{aligned}
		  & \mathcal{A}(Q,K,V)                                      \\
		= & W_qQ+{\rm softmax}(\frac{W_qQ(W_kK)^T}{\sqrt{C}})W_vV ,
	\end{aligned}
	\label{eq:attn}
\end{equation}
where $W_K$, $W_Q$, and $W_V$ are the linear parameters of the attention block. In particular, BPAM utilizes a non-linear composite formulation of $\mathcal{A}$ for co-attention and self-attention operations. This combined formulation can transform the complete pixel-wise attention into prototype-pixel-wise attention, thus decreasing computational expenses accordingly. For prototype co-attention $A_{co}$, we separate the fully pixel-wise query-support attention into proto-support attention and query-proto attention.
\begin{equation}
	\begin{aligned}
		A_{co} = \mathcal{A}(T^{q},P^h,\mathcal{A}(P^h,T^{s},T^{s})),
	\end{aligned}
	\label{eq:pca}
\end{equation}
in which $T^q\in \mathbb{R}^{TH_{l3}W_{l3}\times C}$ and $T^s \in \mathbb{R}^{KH_{l3}W_{l3}\times C}$ are the token of query and support reshaped from feature $\tilde{F}^q_{l3}$ and $\tilde{F}^s_{l3}$, respectively. Similarly, the prototype self-attention $A_{self}$ is computed as:
\begin{equation}
	\begin{aligned}
		A_{self} = \mathcal{A}(T^{q},P^h,\mathcal{A}(P^h,T^{q},T^{q})).
	\end{aligned}
	\label{eq:psa}
\end{equation}

Afterwards, $A_{co}$ and $A_{self}$ are reshaped to $\mathbb{R}^{T\times C\times H_{l3}\times W_{l3}}$, and subsequently concatenated together as holistic attention $A_h$ along the channel dimension:
\begin{equation}
	\begin{aligned}
		A_h = {\rm concat}(A_{co},A_{self}),
	\end{aligned}
	\label{eq:concat_attn}
\end{equation}
$A_h\in \mathbb{R}^{T\times 2C\times H_{l3}\times W_{l3}}$ is then fed to the decoder for predicting query masks.
Therefore, BPAM can explicitly exploit query-to-query relationships for aggregating holistic representation.

\subsection{Training Losses}
We employ the cross-entropy loss $\mathcal{L}_{CE}$ and the IoU loss $\mathcal{L}_{IoU}$ to optimize the sementation mask as follow.
\begin{equation}
	\begin{aligned}
		\mathcal{L}_{CE}=-\frac{\sum_{i=1}^{T}\sum_{j=1}^{HW}({y_{ij}\log{\hat{y}_{ij}}+(1-y_{ij})\log{(1-\hat{y}_{ij})}})}{THW},
	\end{aligned}
	\label{eq:loss_ce}
\end{equation}
\begin{equation}
	\begin{aligned}
		\mathcal{L}_{IoU}=1-\frac{1}{T}\sum_{i=1}^{T}{\frac{\sum_{j=1}^{HW}y_{ij}\hat{y}_{ij}}{\sum_{j=1}^{HW}(y_{ij}+\hat{y}_{ij}-y_{ij}\hat{y}_{ij})}},
	\end{aligned}
	\label{eq:loss_iou}
\end{equation}
where $y_{ij}$ and $\hat{y}_{ij}$ are the corresponding pixel in the predicted mask $\hat{M}^q$ and ground-truth mask $M^q$.

Further, a prototype loss $\mathcal{L}_{proto}$ is proposed to prevent prototypes from over-concentrating to one center during training, ensuring a more balanced distribution of prototypes and thus enhancing the overall performance of our model.

\begin{equation}
	\mathcal{L}_{proto} = \frac{\lambda_{proto}}{N_pK(N_pK-1)}\sum_{i=1}^{N}\sum_{j=1\land j\ne i}^{N}d(p^h_i,p^h_j) ,
	\label{eq:loss_proto}
\end{equation}
where $p^h_i$ is the $i$-th prototype in $P^h$. $\lambda_{proto}$ is the loss coefficient and $d(\cdot,\cdot)$ is the cosine similarity as Eq.~\eqref{eq:cs}. Finally, the total loss $\mathcal{L}_{total}$ is the weighted sum of $\mathcal{L}_{CE}$, $\mathcal{L}_{IoU}$ and $\mathcal{L}_{proto}$ with the cofficients $\lambda_{CE}$, $\lambda_{IoU}$ and $\lambda_{proto}$ respectively.

\begin{table*}[t]
	\caption{Results on YouTube-FSVOS using 5 support images (ResNet-50 is adopted as the backbone), where FT denotes fine-tuning.}
	\centering
	\resizebox{\linewidth}{!}{
		\begin{tabular}{lccccccccccccc}
			\toprule
			\multirow{2}[2]{*}{\textbf{Methods}}  & \multirow{2}[2]{*}{\textbf{Publication}} & \multirow{2}[2]{*}{\textbf{Query}} & \multicolumn{5}{c}{\bm{$\mathcal{F}$}} & \multicolumn{5}{c}{\bm{$\mathcal{J}$}} & \multirow{2}[2]{*}{\textbf{Params}}                                                                                                                         \\
			\cmidrule(lr){4-13}
			                                      &                                          &                                    & 1                                      & 2                                      & 3                                   & 4             & Mean          & 1             & 2             & 3             & 4             & Mean          &       \\
			\midrule
			PMM~\cite{Yang2020PrototypeMM}        & ECCV 2020                                & Image                              & 34.2                                   & 56.2                                   & 49.4                                & 51.6          & 47.9          & 32.9          & 61.1          & 56.81         & 55.91         & 51.7          & 38.6M \\
			PFENet~\cite{Tian2020PriorGF}         & TPAMI 2020                               & Image                              & 33.7                                   & 55.9                                   & 48.7                                & 48.9          & 46.8          & 37.8          & 64.4          & 56.3          & 56.4          & 53.7          & 34.3M \\
			PPNet~\cite{Liu2020PartawarePN}       & TPAMI 2020                               & Image                              & 35.9                                   & 50.7                                   & 47.2                                & 48.4          & 45.6          & 45.5          & 63.8          & 60.4          & 58.9          & 57.1          & 31.5M \\
			RePRI~\cite{Boudiaf2021FewshotSW}     & CVPR 2021                                & Image                              & -                                      & -                                      & -                                   & -             & -             & 45.8          & 68.6          & 59.3          & 64.2          & 59.5          & -     \\
			\midrule
			DANet w/o FT~\cite{Chen2021DelvingDI} & CVPR 2021                                & Video                              & 40.3                                   & 62.3                                   & 60.2                                & 59.4          & 55.6          & 41.5          & 64.8          & 61.3          & 61.4          & 57.2          & 33.3M \\
			DANet~\cite{Chen2021DelvingDI}        & CVPR 2021                                & Video                              & 42.3                                   & 62.6                                   & 60.6                                & 60.0          & 56.3          & 43.2          & 65.0          & 62.0          & 61.8          & 58.0          & 33.3M \\
			TTI w/o DCL~\cite{Siam2022TemporalTI} & arXiv 2022                               & Video                              & -                                      & -                                      & -                                   & -             & -             & 47.5          & 69.5          & 60.5          & 63.8          & 60.3          & -     \\
			TTI~\cite{Siam2022TemporalTI}         & arXiv 2022                               & Video                              & -                                      & -                                      & -                                   & -             & -             & 48.2          & 69.0          & 62.8          & 63.1          & 60.8          & -     \\
			\midrule
			\textbf{HPAN w/o FT}                  & -                                        & Video                              & 47.5                                   & 68.8                                   & 64.3                                & 65.0          & 61.4          & 47.2          & 70.2          & 66.0          & 65.6          & 62.2          & 34.7M \\
			\textbf{HPAN}                         & -                                        & Video                              & \textbf{50.1}                          & \textbf{68.9}                          & \textbf{64.3}                       & \textbf{66.4} & \textbf{62.4} & \textbf{50.2} & \textbf{70.5} & \textbf{66.2} & \textbf{67.0} & \textbf{63.5} & 34.7M \\
			\bottomrule
		\end{tabular}
	}
	\label{tbl:ytvos_cmp}
        \vspace{-10pt}
\end{table*}

\section{Experiments}
\label{sec:experiments}

\subsection{Datasets and Settings}
Following the previous work of DANet~\cite{Chen2021DelvingDI}, we adopt the YouTube-FSVOS dataset~\cite{Chen2021DelvingDI} to evaluate our proposed HPAN. YouTube-FSVOS is built based on YouTube-VOS~\cite{Xu2018YoutubevosAL}, which contains 2,238 videos consisting of 3,774 instances from 40 classes. Compared to FSYTV-40~\cite{fan2022fewshotvo}, another dataset derived from YouTube-VOS with a direct data split of 30 training classes and 10 validation classes, YouTube-FSVOS takes a four-fold cross-validation strategy that each fold contains 10 classes. Specifically,  YouTube-FSVOS leverages 3 folds for training and 1 fold for validation in each experiment and thus enables a more thorough model evaluation across various categories.

Similar to VOS methods~\cite{Perazzi2016ABD, Caelles2017OneShotVO}, we utilize the region similarity in terms of intersection over union ($\mathcal{J}$) and the contour accuracy ($\mathcal{F}$) to validate the performance. All experiments are conducted five times and average results are reported. Following ~\cite{Chen2021DelvingDI} which adopts a $K$-shot setting, $K$ support images (support set) are randomly selected for a specific class $c$ and consecutive frames under the same class are taken as query set. The default value of both the number of support images $K$ and the number of query frames $T$ are set to 5.

\subsection{Implementation Details and Parameters}
Following existing methods, we use ResNet-50~\cite{He2016DeepRL} as our backbone network and are initialized with weights pre-trained on ImageNet-1K\cite{Deng2009ImageNetAL}. The resolution of input images is 241×425 following DANet~\cite{Chen2021DelvingDI} for fair comparison.
During the meta-learning phase, the episodes are sampled from the training set and the parameters of the encoder are frozen. In the fine-tuning phase, we use support images to train the last two layers of the encoder before testing. It is achieved by randomly selecting one test support image as fake fine-tuning query and the remaining as fine-tuning supports. Thus, the encoder with fine-tuning adapts better to the real test query classes during meta-testing.
Adam optimizer is used for HPAN training. We train our model for 40,000 iterations in the meta-learning phase and 100 iterations in the fine-tuning phase. The batch sizes for these two phases are 8 and 1, respectively. The learning rates in these two phases are set to 5e-5 and 2e-5. Moreover, we set the output dimension $C$ of $1\times1$ convolution to 256 in Eq.~\eqref{eq:foreground_l3}.
The scaling coefficients $\lambda_{self}$ and $\lambda_{co}$ in Eq.~\eqref{eq:qga} is set to 0.8 and 0.2, respectively.
As depicted in Eqs.~\eqref{eq:loss_ce}-\eqref{eq:loss_proto}, we train HPAN in the meta-learning phase by relying on three losses, i.e., {\color{black} $\mathcal{L}_{CE}$,  $\mathcal{L}_{IoU}$, $\mathcal{L}_{proto}$}.
The trade-off coefficients for them are 5, 1, and 1, respectively. During fine-tuning, only $\mathcal{L}_{CE}$ is utilized.

\subsection{Comparisons with State-of-the-Arts}
FSVOS is a new task proposed in CVPR-2021 by DANet~\cite{Chen2021DelvingDI}, which the key difference lies in the different query sets. Specifically, by deeming query images in FSS as query video frames in FSVOS, the FSS models can be adapted to the FSVOS task. Therefore, besides FSVOS approaches, we also compare our proposed HPAN with FSS methods. As shown in Table~\ref{tbl:ytvos_cmp}, HPAN has achieved significant improvements on $\mathcal{F}$ and $\mathcal{J}$ in all folds. The mean value of $\mathcal{J}$ for HPAN is 2.7\% better than the current state-of-the-art method TTI~\cite{Siam2022TemporalTI}. Meanwhile, the results of HPAN without fine-tuning are still competitive compared to existing methods, which indicates that the proposed HPAN already contains a good generalization to unseen classes even without fine-tuning. This may pave a new way for future research on FSVOS. Since YouTube-FSVOS shares some common classes with ImageNet, which the backbone is pretrained on, those classes that appear in ImageNet are not `true' unseen ones. Therefore, we compare the knowledge transfer ability of HPAN and DANet~\cite{Chen2021DelvingDI} on the `true' unseen classes that do not appear in ImageNet. Results are shown in Table~\ref{tbl:class_transfer}. In most classes, HPAN performs better in segmenting unseen objects than DANet~\cite{Chen2021DelvingDI}. This is because our HPAN method can fuse all the semantic information of the query set and the support set.

\begin{table}[t]
	\caption{The performance on `true' unseen classes.}
        \vspace{-10pt}
	\begin{center}
		\resizebox{\linewidth}{!}{
			\setlength\tabcolsep{13pt}
			\renewcommand\arraystretch{1.0}
			\begin{tabular}{ccccc}
				\toprule
				\multirow{2}{*}{\textbf{Unseen Class}} & \multicolumn{2}{c}{DANet\cite{Chen2021DelvingDI}} & \multicolumn{2}{c}{\textbf{HPAN}}                                 \\ \cline{2-5}
				~                                      & $\mathcal{F}$                                     & $\mathcal{J}$                     & $\mathcal{F}$ & $\mathcal{J}$ \\
				\midrule
				Deer                                   & 63.6                                              & 66.6                              & \textbf{71.0} & \textbf{72.3} \\
				Giraffe                                & 73.2                                              & 70.2                              & \textbf{79.8} & \textbf{74.9} \\
				Hand                                   & \textbf{50.5}                                     & \textbf{57.0}                     & 49.5          & 54.5          \\
				Parrot                                 & 59.3                                              & 65.0                              & \textbf{71.0} & \textbf{75.2} \\
				Person                                 & 36.2                                              & 29.4                              & \textbf{39.1} & \textbf{32.4} \\
				Skateboard                             & 36.8                                              & 16.3                              & \textbf{46.8} & \textbf{53.3} \\
				Surfboard                              & 19.2                                              & \textbf{23.0}                     & \textbf{35.4} & 16.0          \\
				Tennis racket                          & 23.7                                              & 14.4                              & \textbf{25.6} & \textbf{17.2} \\
				\midrule
				Mean                                   & 49.1                                              & 42.7                              & \textbf{52.3} & \textbf{49.5} \\
				\bottomrule
			\end{tabular}
		}
	\end{center}
	\label{tbl:class_transfer}
    \vspace{-10pt}
\end{table}

\begin{table}[t]
	\caption{Effects of prototype loss coefficient in HPAN.}
        \vspace{-10pt}
	\begin{center}
		\resizebox{\linewidth}{!}{
			\setlength\tabcolsep{13pt}
			\renewcommand\arraystretch{1.0}
			\begin{tabular}{ccccc}
				\toprule
				\multirow{2}[2]{*}{\bm{$\lambda_{proto}$}} & \multicolumn{2}{c}{\textbf{w/o fine-tuning}} & \multicolumn{2}{c}{\textbf{w/ fine-tuning}}                                 \\
				\cmidrule(lr){2-5}
				                                           & $\mathcal{F}$                                & $\mathcal{J}$                               & $\mathcal{F}$ & $\mathcal{J}$ \\
				\midrule
				0                                          & 61.2                                         & 61.6                                        & 62.3          & 63.0          \\
				1                                          & \textbf{61.4}                                & \textbf{62.2}                               & \textbf{62.4} & \textbf{63.5} \\
				2                                          & 60.6                                         & 62.0                                        & 61.5          & 63.1          \\
				\bottomrule
			\end{tabular}}
	\end{center}
	\label{tbl:loss_abs}
    \vspace{-10pt}
\end{table}

\begin{table}[!t]
	\caption{Effects of prototype numbers in HPAN.}
        \vspace{-10pt}
	\begin{center}
		\resizebox{\linewidth}{!}{
			\setlength\tabcolsep{8pt}
			\renewcommand\arraystretch{1.0}
			\begin{tabular}{ccccc}
				\toprule
				\textbf{Prototype Number} & \multicolumn{2}{c}{\textbf{w/o fine-tuning}} & \multicolumn{2}{c}{\textbf{w/ fine-tuning}}                                 \\
				\cmidrule(lr){2-5}
				$N_p$                     & $\mathcal{F}$                                & $\mathcal{J}$                               & $\mathcal{F}$ & $\mathcal{J}$ \\
				\midrule
				1                         & 61.3                                         & 61.9                                        & 62.2          & \textbf{63.5} \\
				5                         & \textbf{61.4}                                & \textbf{62.2}                               & \textbf{62.4} & \textbf{63.5} \\
				10                        & 61.0                                         & 61.8                                        & 62.1          & 63.0          \\
				15                        & 60.0                                         & 60.9                                        & 61.5          & 62.6          \\
				\bottomrule
			\end{tabular}}
	\end{center}
	\label{tbl:proto_abs}
    \vspace{-10pt}
\end{table}

\subsection{Ablation Study}

\textbf{Effects of Prototype Loss Coefficient.}
By further varying the value of $\lambda_{proto}$ in \{0,1,2\}, we show the results of the mean $\mathcal{J}$ and $\mathcal{F}$ for HPAN in Table~\ref{tbl:loss_abs}. The prototype loss function is employed to counteract prototype over-concentration, promoting a balanced prototype distribution across instances and representing local target features better. If the coefficient $\lambda_{proto}$ is set to $0$, the k-means algorithm may excessively concentrate the prototypes to one center, leading to suboptimal segmentation of unseen class target details. Conversely, when setting $\lambda_{proto} \ge 2$, the strong constraints of loss may force prototypes outside the target features' distribution range, potentially degrading segmentation quality. Table~\ref{tbl:loss_abs} demonstrates that setting $\lambda_{proto} = 1$ effectively balances the distribution of prototypes within the feature space, producing superior results compared to $\lambda_{proto}$ values of 0 and 2.

\textbf{Effects of Prototype Number.}
We vary the value of $N_p$ from \{1,5,10,15\} to analyze the mean $\mathcal{J}$ and $\mathcal{F}$ results when the backbone is / is not fine-tuned.
As revealed in Table~\ref{tbl:proto_abs}, in most cases, a smaller $N_p$ leads to better $\mathcal{J}$ and $\mathcal{F}$.
When $N_p=5$, we obtain the best $\mathcal{J}$ and $\mathcal{F}$ under the setting of w/o fine-tuning.
Meanwhile, both $N_p=1$ and $N_p=5$ can achieve desirable results under the setting of fine-tuning.
Accordingly, we set the default value of $N_p$ to 5 for achieving satisfactory results.
Results with fine-tuning in Table~\ref{tbl:proto_abs} are actually better than the ones w/o fine-tuning.

\begin{table}[t]
	\caption{Component analysis on our HPAN model.}
        \vspace{-10pt}
	\begin{center}
		\resizebox{\linewidth}{!}{
			\setlength\tabcolsep{6pt}
			\renewcommand\arraystretch{1.0}
			\begin{tabular}{ccccccc}
				\toprule
				\multirow{2}{*}{\textbf{Methods}} & \multirow{2}{*}{\textbf{PGAM}} & \multirow{2}{*}{\textbf{BPAM}} & \multicolumn{2}{c}{\textbf{w/o fine-tuning}} & \multicolumn{2}{c}{\textbf{w/ fine-tuning}}                                 \\
				\cmidrule(lr){4-7}
				                                  &                                &                                & $\mathcal{F}$                                & $\mathcal{J}$                               & $\mathcal{F}$ & $\mathcal{J}$ \\
				\midrule
				Baseline                          &                                &                                & 55.2                                         & 56.0                                        & 55.8          & 57.2          \\
				\cmidrule(lr){1-7}
				\multirow{3}{*}{\textbf{HPAN}}
				                                  &                                & $\surd$                        & 58.6                                         & 58.9                                        & 59.4          & 60.4          \\
				                                  & $\surd$                        &                                & 59.0                                         & 59.4                                        & 60.3          & 61.8          \\
				                                  & $\surd$                        & $\surd$                        & \textbf{61.4}                                & \textbf{62.2}                               & \textbf{62.4} & \textbf{63.5} \\
				\bottomrule
			\end{tabular}}
	\end{center}
	\label{tbl:mudule_abs}
    \vspace{-10pt}
\end{table}

\begin{table}[t]
	\caption{Comparison of cost time (ms), and FPS when different numbers of support images $K$ and query frames $T$.}
        \vspace{-10pt}
	\begin{center}
		\resizebox{\linewidth}{!}{
			\setlength\tabcolsep{15pt}
			\renewcommand\arraystretch{1.0}
			\begin{tabular}{cccccc}
				\toprule
				\multicolumn{2}{c}{\textbf{Method}} & \multicolumn{2}{c}{DANet\cite{Chen2021DelvingDI}} & \multicolumn{2}{c}{\textbf{HPAN}}                                    \\
				\midrule
				$K$                                 & $T$                                               & Time                              & FPS & Time         & FPS         \\
				\midrule
				5                                   & 5                                                 & 103                               & 49  & \textbf{91}  & \textbf{55} \\
				5                                   & 10                                                & 156                               & 64  & \textbf{141} & \textbf{71} \\
				10                                  & 10                                                & 202                               & 49  & \textbf{175} & \textbf{57} \\
				10                                  & 20                                                & 304                               & 66  & \textbf{275} & \textbf{73} \\
				20                                  & 20                                                & 401                               & 50  & \textbf{343} & \textbf{58} \\
				20                                  & 40                                                & 618                               & 65  & \textbf{570} & \textbf{70} \\
				40                                  & 40                                                & 804                               & 50  & \textbf{775} & \textbf{52} \\
				\bottomrule
			\end{tabular}}
	\end{center}
	\label{tbl:time_Nq_Ns}
    \vspace{-10pt}
\end{table}

\begin{table}[!t]
	\caption{Comparison of cost time (ms) and FPS when different numbers of prototypes $N_p$.}
        \vspace{-10pt}
	\begin{center}
		\resizebox{\linewidth}{!}{
			\setlength\tabcolsep{17pt}
			\renewcommand\arraystretch{1.0}
			\begin{tabular}{ccccc}
				\toprule
				\textbf{Method}                & \bm{$N_p$} & \textbf{Time} & \textbf{FPS} \\
				\midrule
				DANet\cite{Chen2021DelvingDI}  & -          & 103           & 49           \\
				\midrule
				\multirow{5}{*}{\textbf{HPAN}} & 1          & 65            & 77           \\
				                               & 5          & 91            & 55           \\
				                               & 10         & 120           & 42           \\
				                               & 15         & 157           & 32           \\
				                               & 20         & 183           & 27           \\
				\bottomrule
			\end{tabular}}
	\end{center}
	\label{tbl:time_Np}
    \vspace{-10pt}
\end{table}

\textbf{Component Analysis.}
The key components of HPAN are the prototype graph attention module (PGAM) and the bidirectional prototype attention module (BPAM).
Hence, we perform component analysis on PGAM and BPAM using the baseline model and our HPAN model.
Here, we use a simple DANet~\cite{Chen2021DelvingDI} without PGAM and BPAM for baseline development, which uses the prototypes clustered only from the middle frame and disable the prototypes self-attention.
Results are present in Table~\ref{tbl:mudule_abs}.
From this table, it is evident that 1) PGAM can significantly improve performance and 2) additional gains are achieved by BPAM.

\textbf{Computation Cost Analysis.}
We analyze the running efficiency by comparing HPAN with DANet~\cite{Chen2021DelvingDI}.
We conduct this test on a V100-16G GPU to guarantee consistent model size and training environment.
In Table~\ref{tbl:time_Nq_Ns}, we find that both DANet~\cite{Chen2021DelvingDI} and our HPAN take more time when increasing linearly the number of support images $K$ and query frames $T$ when the number of prototypes $N_p=5$.
When the ratio of K and T is constant, the FPS of the two algorithms remains stable.
From Table~\ref{tbl:time_Np}, we can notice that the cost time of HPAN increases significantly as the number of prototypes $N_p$ increases when $K=5$ and $T=5$.
HPAN shows higher running efficiency than DANet~\cite{Chen2021DelvingDI} when the number of prototypes $N_p \le 5$ and becomes slower than the comparison method when $N_p \ge 10$.

\begin{table}[t]
	\caption{Performance comparison for different numbers of query frames $T$.}
        \vspace{-10pt}
	\begin{center}
		\resizebox{\linewidth}{!}{
			\setlength\tabcolsep{17pt}
			\renewcommand\arraystretch{1.0}
			\begin{tabular}{ccccc}
				\toprule
				\multirow{2}{*}{\bm{$T$}} & \multicolumn{2}{c}{\textbf{w/o fine-tuning}} & \multicolumn{2}{c}{\textbf{with fine-tuning}}                                 \\
				\cmidrule(lr){2-5}
				                          & $\mathcal{F}$                                & $\mathcal{J}$                                 & $\mathcal{F}$ & $\mathcal{J}$ \\
				\midrule
				1                         & 63.3                                         & 64.6                                          & 63.8          & 65.4          \\
				2                         & 62.5                                         & 63.4                                          & 63.0          & 64.2          \\
				5                         & 61.4                                         & 62.2                                          & 62.4          & 63.5          \\
				10                        & 61.1                                         & 61.9                                          & 62.2          & 63.4          \\
				\bottomrule
			\end{tabular}
		}
	\end{center}
	\label{tbl:t_performance}
        \vspace{-10pt}
\end{table}

\begin{table}[t]
	\caption{Performance comparison for different numbers of support images $K$.}
        \vspace{-10pt}
	\begin{center}
		\resizebox{\linewidth}{!}{
			\setlength\tabcolsep{17pt}
			\renewcommand\arraystretch{1.0}
			\begin{tabular}{ccccc}
				\toprule
				\multirow{2}{*}{\bm{$K$}} & \multicolumn{2}{c}{\textbf{w/o fine-tuning}} & \multicolumn{2}{c}{\textbf{with fine-tuning}}                                 \\
				\cmidrule(lr){2-5}
				                          & $\mathcal{F}$                                & $\mathcal{J}$                                 & $\mathcal{F}$ & $\mathcal{J}$ \\
				\midrule
				1                         & 58.4                                         & 58.7                                          & 58.4          & 58.9          \\
				2                         & 59.6                                         & 60.1                                          & 60.3          & 60.9          \\
				5                         & 61.4                                         & 62.2                                          & 62.4          & 63.5          \\
				10                        & 62.2                                         & 62.8                                          & 63.7          & 64.7          \\
				\bottomrule
			\end{tabular}
		}
	\end{center}
	\label{tbl:k_performance}
        \vspace{-10pt}
\end{table}

\begin{figure*}[t]
	\centering
	\includegraphics[width=1\linewidth]{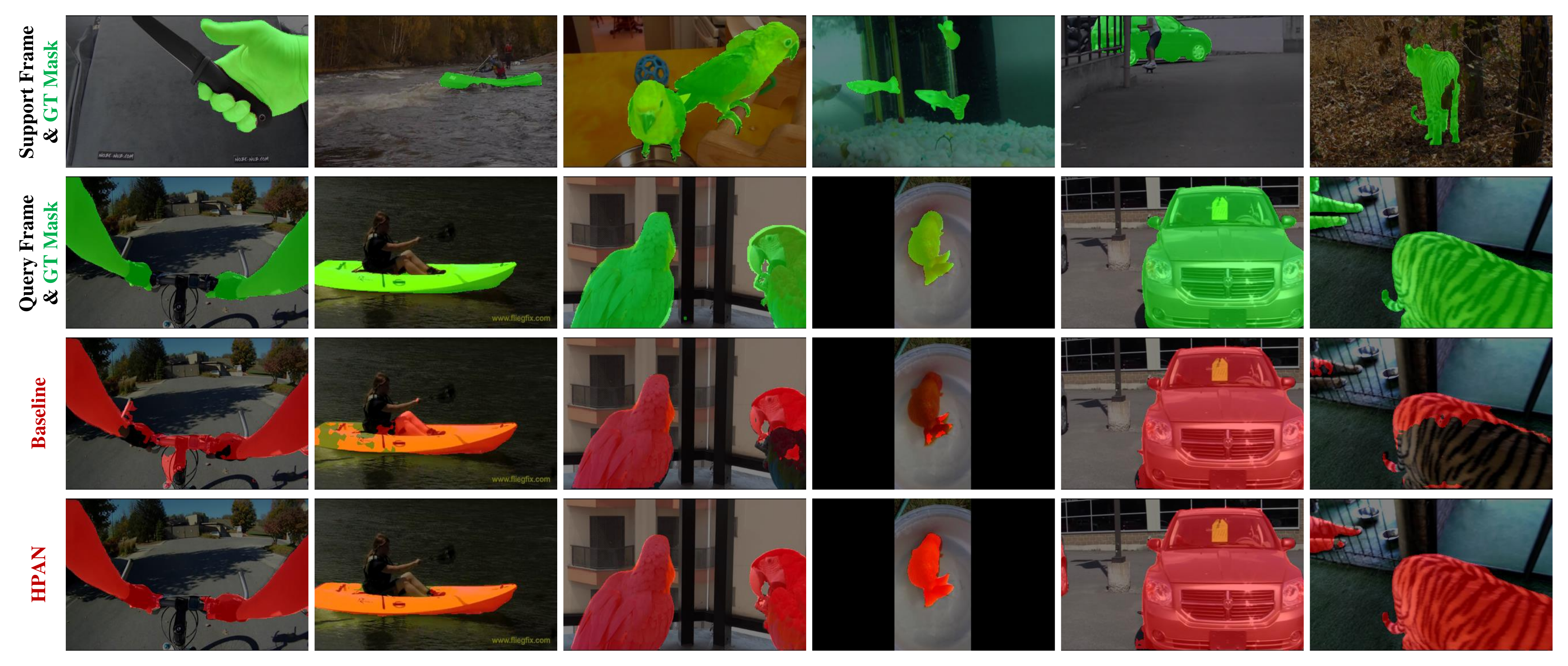}
	\caption{Segmentation visualizations of HPAN and baseline models on unseen image frames. Each column shows one test frame and its corresponding prediction. From top to bottom, each row indicates support images along with their ground-truth masks (\textcolor{green}{green}), query frames (for testing) along with their ground-truth masks (\textcolor{green}{green}), baseline predictions (\textcolor{red}{red}), and HPAN predictions (\textcolor{red}{red}), respectively.}
	\label{fig:vis_imgs}
        \vspace{-10pt}
\end{figure*}

\textbf{Effects of Support and Query Number.}
In Table~\ref{tbl:t_performance}, we show the segmentation performance with and without fine-tuning for different numbers of query frames $T$. Both the $\mathcal{F}$ and $\mathcal{J}$ metrics of the HPAN will decrease when the number of query frames increases.
For different numbers of support images $K$, Table~\ref{tbl:k_performance} shows that the performance of both $\mathcal{F}$ and $\mathcal{J}$ will be improved by providing more support images.

\begin{figure}[t]
	\centering
	\includegraphics[width=1\linewidth]{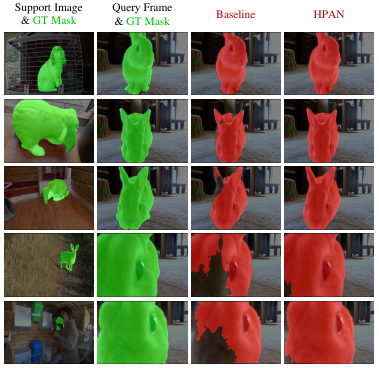}
	\caption{Segmentation visualizations of HPAN and baseline models on an entire episode with five support images and five query frames.}
	\label{fig:vis_video}
        \vspace{-10pt}
\end{figure}

\begin{figure*}[t]
	\centering
	\includegraphics[width=0.98\linewidth]{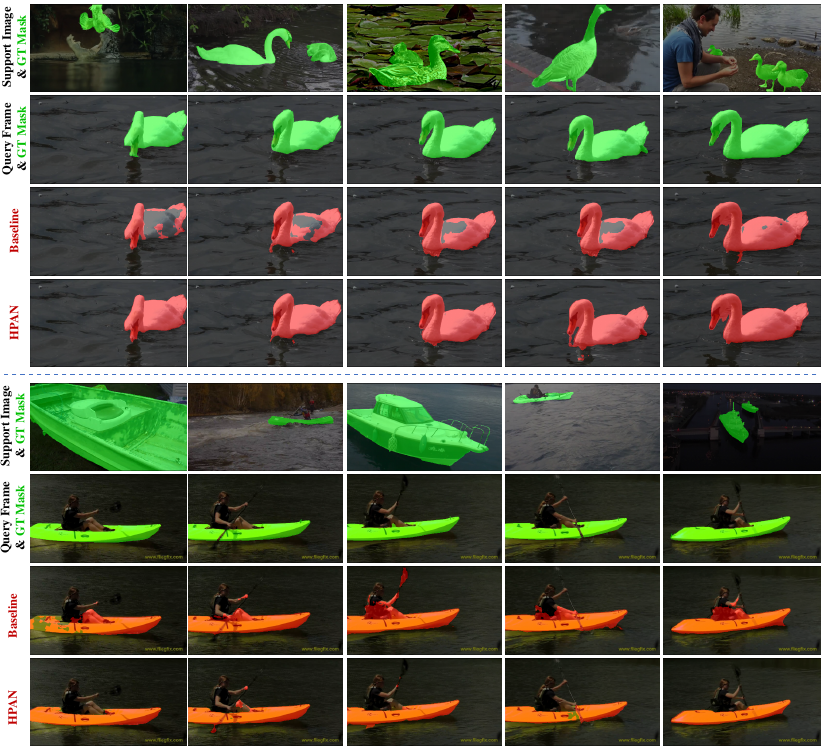}
	\caption{
		Segmentation visualizations of HPAN and baseline models on unseen image frames. From top to bottom, each row indicates support images and their ground-truth (GT) masks (\textcolor{green}{green}), query frames (for testing) and their GT masks (\textcolor{green}{green}), baseline predictions (\textcolor{red}{red}), and HPAN predictions (\textcolor{red}{red}), respectively.
	}
	\label{fig:visualizations}
        \vspace{-10pt}
\end{figure*}

\subsection{Qualitative Results}
\begin{figure}[t]
	\centering
	\includegraphics[width=0.99\linewidth]{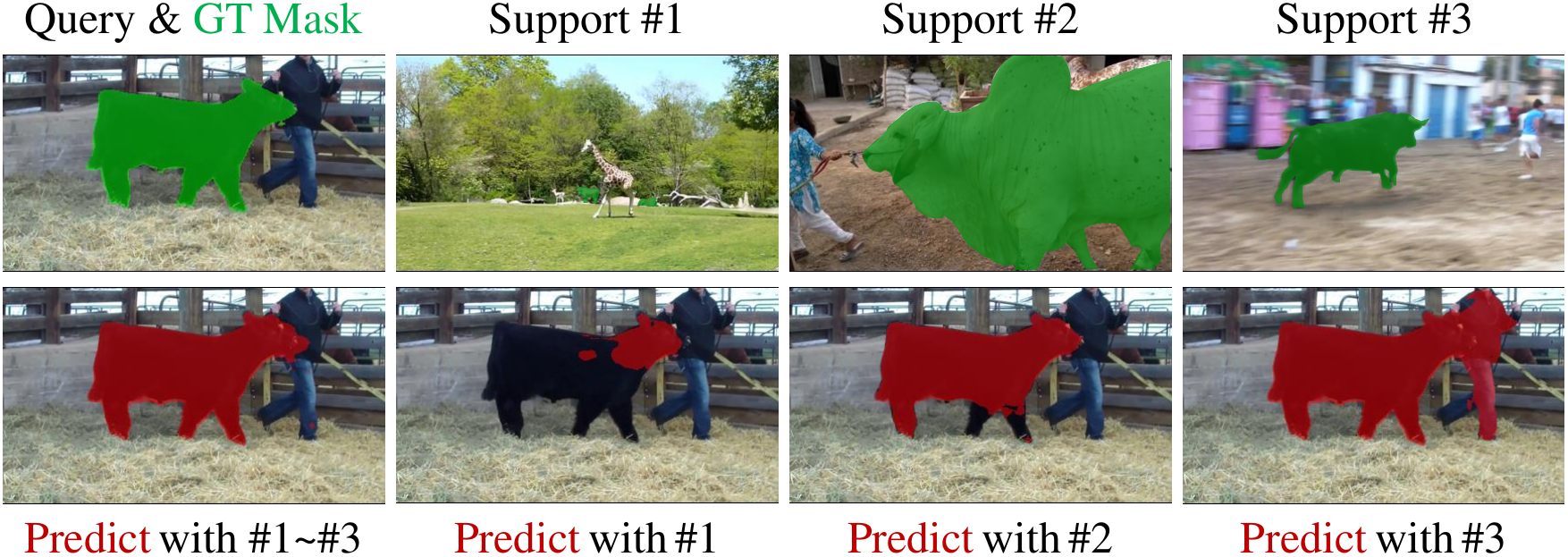}
	\caption{Effects of different quality support images on the same query frame.}
	\label{fig:support_quality}
    \vspace{-10pt}
\end{figure}

\begin{figure}[t]
	\centering
	\includegraphics[width=1\linewidth]{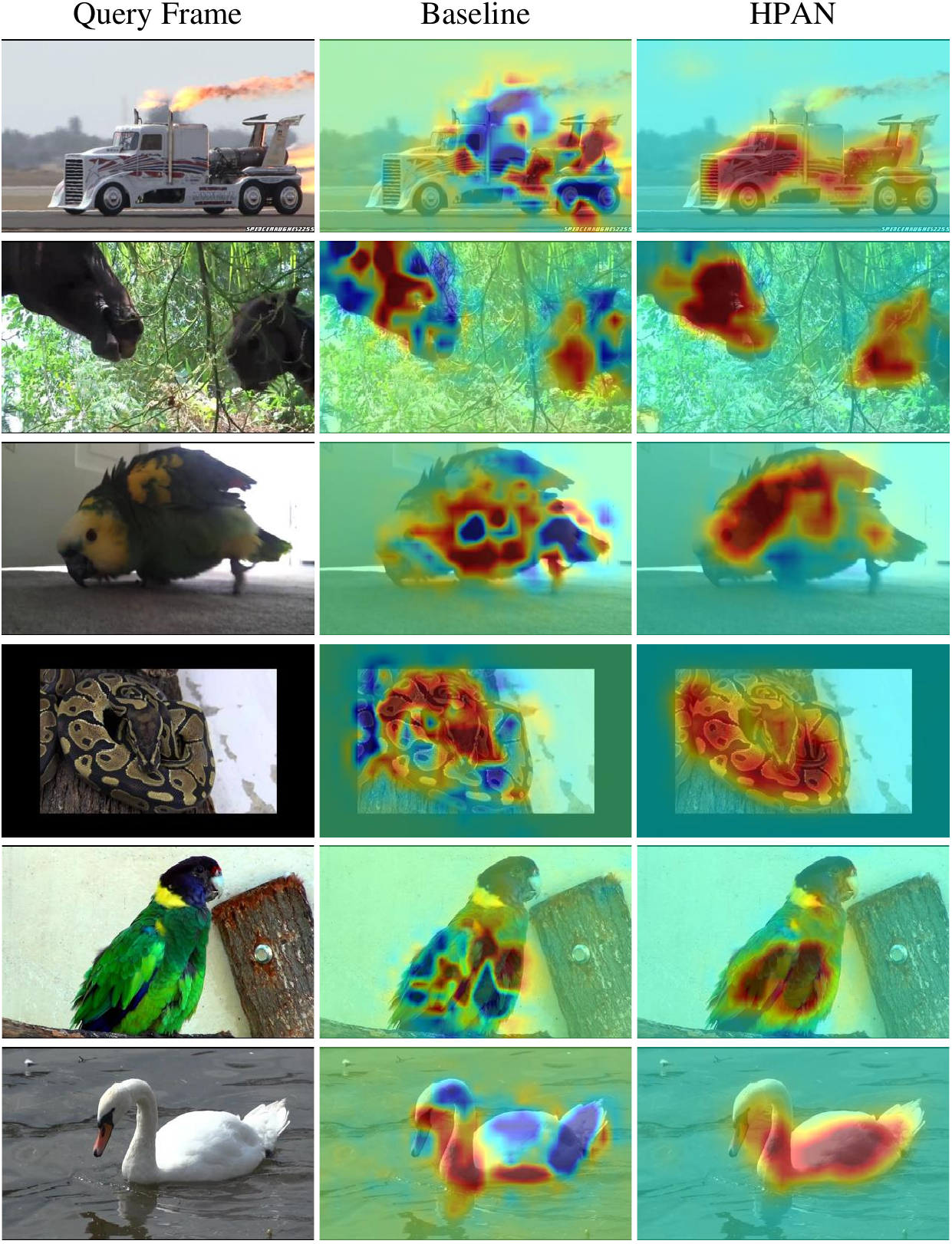}
	\caption{Attention visualization of HPAN and baseline models on unseen image frames. From left to right, each column represents the query frame, baseline attention map, and HPAN attention map. The red and blue regions in the attention map denote positive and negative activations respectively.
	}
	\label{fig:attention_cam}
    \vspace{-10pt}
\end{figure}

We randomly select test frames to visualize their object segmentation masks. The qualitative comparisons between HPAN and baseline (without PGAM and BPAM) are depicted in Fig.~\ref{fig:vis_imgs}. As demonstrated in this figure, HPAN can capture more holistic and accurate foreground objects. The predicted masks are well matched with the ground-truth masks in most cases (\eg, {\it bird}, {\it fish}, and {\it tiger} in the \#2, \#3, and \#6 columns of Fig.~\ref{fig:vis_imgs}). Further, in columns \#1 and \#2, HPAN avoids segmenting the bicycle handlebar and the oarsman in the background while the baseline fails, showing its ability to suppress background noise. In columns \#5 and \#6, HPAN segments occluded vehicle and tiger based on the wheel and the claw located at the edge of the frame, which indicates that the prototypes extracted by our method can better capture local features. Fig.~\ref{fig:vis_video} shows the tracking performance of HPAN in a scale-changing scenario containing an approaching rabbit. We show more qualitative results on video clips in Fig.~\ref{fig:visualizations} to demonstrate the temporal consistency of the output masks. Compared to the baseline, our method can better segment unseen objects by maintaining inter-frame temporal consistency. Fig.~\ref{fig:support_quality} indicates that HPAN can improve segmentation results by varying qualities of support images. As for comparing the qualities of the attention maps on HPAN and baseline without GPAM and BPAM, we generated the Grad-CAM~\cite{selvaraju2017grad} for the attention map $A_h$ in Fig.~\ref{fig:attention_cam}. Due to suppressing the background, HPAN reduces the negative activation of foreground objects and focuses more on local regions. Our method can segment objects completely and filter background noise compared to baseline methods.

\section{Limitation and Social Impact}
\subsection{Limitation and Future Work}
Our method employs the k-means algorithm for clustering local prototypes effectively. However, a limitation of the k-means algorithm is its inability to adjust the distribution of cluster centers. To address this, we propose a prototype loss function to prevent the over-concentration of cluster centers. Nevertheless, this solution requires a balance between the dispersion and centralization of prototypes by manually tuning the hyperparameters. The Sinkhorn-Knopp algorithm~\cite{knight2008thesa}, as evidenced by its efficacy in existing methods~\cite{wang2023visualrw, Zhou2022RethinkingSS}, optimizes the matching distribution between cluster centers and features. This algorithm inherently inhibits subclusters from excessive concentration or total rejection, potentially obviating the need for a prototype loss function and enabling a more efficient clustering process. In future work, we intend to explore the efficacy of such alternative clustering algorithms.

When considering the robustness of our HPAN to adversarial attacks~\cite{wei2022physicalaa, cheng2022physicalao} in real-world applications, it is a growing concern within the community.
To counter these threats, several adversarial defense strategies have been put forward, such as image restoration~\cite{gupta2019ciidefenceda} and adversarial training~\cite{pandey2022adversariallyrp}, all aimed at bolstering the defense capabilities of the network.
Therefore, exploring adversarial attacks and defenses for FSVOS in practical scenarios has emerged as a significant area of research, highlighting our future potential improvement points.

\subsection{Social Impact}
The proposed HPAN can be applied in many real-world scenarios, such as multimedia editing, video surveillance, and autonomous driving. It will help humans quickly segment some unseen objects in the video, although it cannot reach human cognitive abilities. We must emphasize that it is crucial for our method to preserve surveillance privacy and strengthen security driving measures when deployed in some risky scenes.

\section{Conclusion}
\label{sec:conclusion}

In this paper, we proposed a holistic prototype attention network (HPAN) to address the challenging few-shot video object segmentation task by leveraging holistic knowledge. Specifically, our proposed prototype graph attention module generated representative prototypes among all foreground features. Then, the proposed bidirectional prototype attention module combined co-attention and self-attention to infer unseen objects in a transductive manner. Comprehensive Experimental results on the YouTube-FSVOS dataset confirmed that HPAN achieved state-of-the-art performance compared with existing methods.

\bibliographystyle{IEEEtran}
\bibliography{ref}

\end{document}